\documentclass[10pt,twocolumn,letterpaper]{article}
\usepackage{pifont}
\newcommand{\cmark}{\textcolor{green!50!black}{\ding{51}}}
\newcommand{\xmark}{\textcolor{red!70!black}{\ding{55}}}
\usepackage[pagenumbers]{wacv} % To force page numbers, e.g. for an arXiv version

\usepackage{float}
\usepackage{placeins}
\usepackage{svg}

\definecolor{wacvblue}{rgb}{0.21,0.49,0.74}
\usepackage[pagebackref,breaklinks,colorlinks,allcolors=wacvblue]{hyperref}
\usepackage{multirow}

\def\wacvPaperID{2861} % *** Enter the WACV Paper ID here
\def\confName{WACV}
\def\confYear{2027}

\title{MuyBridge: Mobile Human Center-of-Mass Estimation from Monocular Video via Sparse Fusion}

\author{
Aidan Bradshaw$^{1}$, Marco Giordano$^{1}$, David Rode$^{1}$,
Andreas Habersack$^{2}$, Elif Basokur$^{1}$, Annika Kruse$^{2}$,\\
Markus Tilp$^{2}$, Michele Magno$^{1}$, Peter Wolf$^{1}$,
Luca Benini$^{1,3}$, Christoph Leitner$^{1}$\\[4pt]
{
$^{1}$ETH Zurich \qquad
$^{2}$University of Graz \qquad
$^{3}$University of Bologna}
}

\begin{document}
\maketitle

\begin{abstract}
The 3D center of mass (CoM) is a primary quantity in the biomechanical analysis of sport, rehabilitation, and clinical movement, yet existing 3D pose tracking, mesh recovery, and multi-view triangulation methods either optimize 3D keypoint accuracy without anatomical constraints or carry compute and capture infrastructure too heavy to deploy where CoM tracking is most useful. As a result, the metric CoM remains difficult for coaches and movement analysts to measure from a single camera where athletes train and compete. In this work, we introduce MuyBridge, an on-device system that estimates the athlete's segmental center of mass trajectory from a single phone camera video stream. MuyBridge couples a compact 2D pose network and a distilled single-step monocular depth network through an analytic metric fusion that uses anatomical and physical priors to anchor the metric CoM, requiring no 3D or task-specific supervision. Evaluated on the athletic movements of AthletePose3D (running, track and field, and figure skating), MuyBridge achieves 33–41 mm vertical CoM error and 2.3–6.6\% absolute-relative range error (AbsRel) under a one-time calibration, and produces CoM estimates at the 63~FPS pose-estimation rate using asynchronous 2.86 Hz depth updates on iPhone 15. Code is available at:~\url{https://github.com/Abradshaw1/Muybridge}.
\end{abstract}

\section{Introduction}
The whole-body center of mass (CoM) is a widely used quantity in segmental biomechanics for analyzing balance, acceleration, and landing in athletic performance~\cite{winter1995balance,mapelli2014sportcom,lin2014comacceleration,bates2013landing}. Estimating it, however, commonly requires metric body-segment positions together with segment lengths, mass fractions, and segment-specific centers of mass, obtained from marker-based motion capture~\cite{mapelli2014sportcom,forsell2009markers,halvorsen2009runningcom}, force plates~\cite{shimba1984forceplatform,barbier2003forceplate,gutierrezfarewik2006com}, or wearable inertial sensing~\cite{labrozzi2024imucom,simonetti2021imucom}. This often confines athletes to laboratories and dedicated capture facilities rather than the fields, arenas, and training environments where they normally train, making it difficult to measure their natural movement during competition.

Computer vision offers a way to measure similar movements with substantially
less equipment. Two-dimensional pose estimators recover image-space joint
locations~\cite{sun2019hrnet,xu2024vitposepp,khirodkar2024sapiens},
2D-to-3D lifting methods recover 3D joint configurations~\cite{zhu2023motionbert,mehraban2024motionagformer,huang2025posemamba},
mesh-recovery methods estimate body pose and shape~\cite{kanazawa2018hmr,goel2023humans4d,patel2025camerahmr},
and monocular depth models estimate scene geometry~\cite{midas2020,hu2024metric3dv2,piccinelli2026unidepthv2}.
Although useful, these methods are generally optimized for keypoint, pose, surface, or depth accuracy and do not directly recover anatomical quantities. A smaller body of work incorporates biomechanical structure through musculoskeletal models~\cite{uhlrich2023opencap,xia2025hsmr} or mass and contact constraints~\cite{Tripathi_2023_CVPR}, but often relies on computationally heavy 3D body representations~\cite{xia2025hsmr,koleini2025biopose,kaichi2018com} or calibrated multi-view capture and cloud processing~\cite{uhlrich2023opencap}, limiting their use and practicality during routine athletic measurement. This leaves segmental CoM monitoring on deployable platforms largely unexplored.

We introduce MuyBridge, a monocular system for estimating metric segmental CoM from phone video. MuyBridge combines a compact pose network that identifies image-space anatomical keypoints defining the athlete’s body segments with monocular depth sampled at these locations to capture relative range and track camera-to-athlete motion. The system then uses an analytic fusion of the pose and depth predictions with stature-scaled anthropometric constraints to recover metric range, while a one-time scene calibration supplies the ground-plane geometry required by complementary physical cues. The resulting fused range places the keypoints as metric 3D segment endpoints. MuyBridge maps these endpoints to segment geometries, mass fractions, and longitudinal CoM locations using stature-scaled, sex-specific de Leva parameters~\cite{deleva1996adjusted}, then computes whole-body CoM as a mass-weighted sum of the resulting segment centers.

We evaluate MuyBridge on AthletePose3D~\cite{yeung2025athletepose3d} across running, track and field, and figure skating. Across these motion regimes, MuyBridge achieves 33--41~mm vertical CoM error and 2.3--6.6\% absolute-relative range error (AbsRel). We evaluate aggregate performance and characterize how viewpoint, motion phase, and individual range cues affect monocular CoM recovery. On an iPhone~15, the pose network runs at 63~FPS while asynchronous depth updates at 2.86~Hz support CoM estimation at the pose rate. 

\begin{itemize}
\item MuyBridge, a deployable single-camera method for metric segmental CoM estimation that combines image-space pose, subject-specific anthropometry, sparse monocular depth, and geometric range constraints without task-specific 3D human-pose or CoM supervision.
\item An evaluation across cyclic, rotational, and ballistic athletic motions that characterizes where MuyBridge performs well and where camera-to-athlete range recovery becomes the dominant source of error, including the effects of viewpoint, motion phase, and individual range cues.
\item A hardware-aware mobile implementation that compresses and quantizes the pose and depth networks, executes them asynchronously, and profiles latency, throughput, memory, and energy on an iPhone~15.
\end{itemize}

\begin{figure*}[t]
\centering
\includegraphics[width=\linewidth]{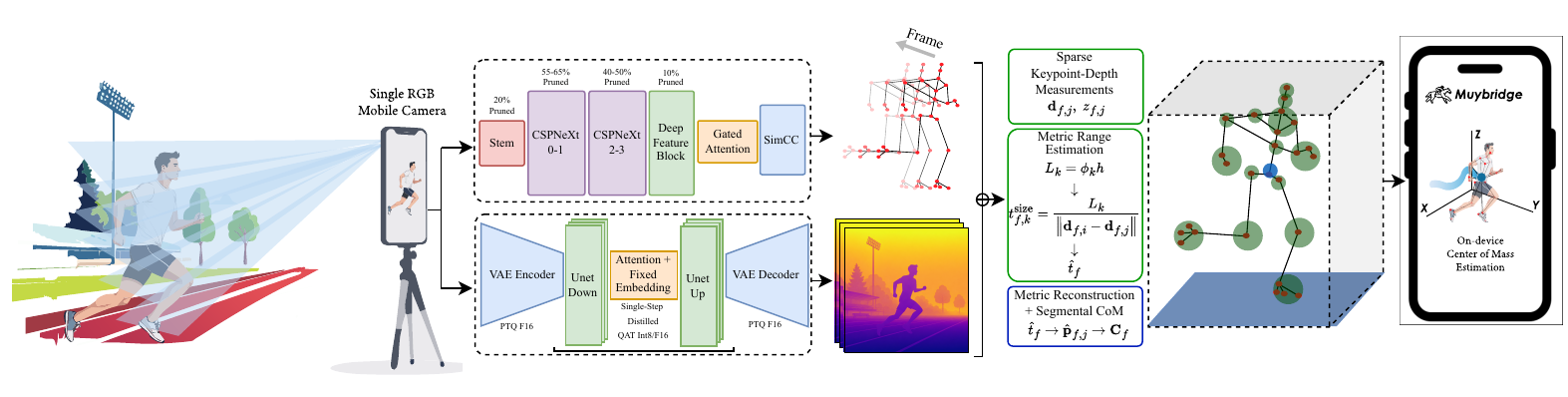}
\caption{Overview of MuyBridge. A single RGB phone camera provides input to the compact pose and monocular depth networks, whose outputs are combined through metric fusion to recover metric joint positions and compute whole-body CoM on device.}
\label{fig:pipeline}
\end{figure*}

\section{Related Work}
\noindent\textbf{Biomechanically grounded vision.}
Recent work has moved beyond keypoint accuracy toward anatomically and
physically constrained human-motion estimation. BioPose
~\cite{koleini2025biopose} incorporates biomechanical structure into monocular
3D pose estimation, while HSMR~\cite{xia2025hsmr} reconstructs humans using a
biomechanically accurate skeletal model. IPMAN~\cite{Tripathi_2023_CVPR}
introduces center-of-mass, center-of-pressure, and body--floor constraints into
monocular human reconstruction, and OpenCap~\cite{uhlrich2023opencap} evaluates pose estimator through the musculoskeletal models. Kaichi et al.~\cite{kaichi2018com}
estimate CoM directly from multi-view visual-hull reconstruction. These works
show the value of anatomical and physical structure, but metric biomechanical
reconstruction commonly relies on richer body models, optimization, or
multi-view capture, rather than targeting deployable platforms such as mobile hardware.

\noindent\textbf{Monocular geometry and scale.}
Recent lifting methods improve root-relative 3D pose through kinematic,
temporal, and state-space representations~\cite{peng2024ktpformer,cui2026sasmamba}, while mesh-recovery methods such as
PromptHMR~\cite{wang2025prompthmr}, SAM 3D Body~\cite{yang2026sam3dbody}, and Neural Localizer Fields~\cite{sarandi2024nlf} recover increasingly detailed body geometry.
Estimating articulated body configuration, however, does not necessarily
determine absolute camera-to-subject range. RootNet~\cite{moon2019rootnet} explicitly estimates camera distance, metric-depth
methods recover scene scale~\cite{bhat2023zoedepth,bochkovskii2025depthpro,
wang2025moge2,piccinelli2024unidepth,hu2024metric3dv2}, and
affine-invariant approaches such as Marigold~\cite{ke2025marigold}
recover relative scene geometry without fixing metric scale. These approaches
address complementary components of monocular 3D reconstruction, but do not
themselves define the metric segmental quantities required for whole-body
CoM estimation.
 
\noindent\textbf{Efficient visual inference.}
Human pose estimation has increasingly targeted lower-cost inference alongside
accuracy. RTMPose~\cite{jiang2023rtmpose} provides a compact 2D pose architecture, while recent 2D-to-3D methods reduce the cost of temporal pose modeling through compact transformer and state-space representations~\cite{mehraban2024motionagformer,huang2025posemamba,cui2026sasmamba}. Monocular depth models have similarly reduced the cost of scene geometry estimation, showing feasibility for efficient human modeling under resource constraints.

\section{Method}
\label{sec:method}
\subsection{Preliminaries}
\label{sec:basis}
\noindent\textbf{Segmental Center-of-Mass Model.}
We compute whole-body CoM using the segmental anthropometric model of Zatsiorsky and Seluyanov, as re-expressed by de Leva~\cite{zatsiorsky1990estimation,deleva1996adjusted}, representing the body as rigid segments whose endpoints are anatomical joint centers. For segment $s$ with proximal and distal endpoints $\mathbf{p}^{\mathrm{prox}}_s$ and $\mathbf{p}^{\mathrm{dist}}_s$, the segment center of mass is

\begin{equation}
\mathbf{c}_s =
\mathbf{p}^{\mathrm{prox}}_s +
\rho_s
\left(
\mathbf{p}^{\mathrm{dist}}_s -
\mathbf{p}^{\mathrm{prox}}_s
\right),
\label{eq:segcom}
\end{equation}
where $\rho_s$ specifies the longitudinal position of the center of mass along segment $s$. Whole-body CoM is then computed as the mass-weighted sum of the segment centers,

\begin{equation}
\mathbf{C} =
\frac{\sum_s m_s \mathbf{c}_s}
{\sum_s m_s},
\label{eq:bodycom}
\end{equation}
where $m_s$ is the mass fraction of segment $s$. We use sex-specific segment mass fractions and longitudinal CoM positions from this formulation. Computing $\mathbf{C}$ therefore requires the metric locations of the anatomical segment endpoints together with the corresponding anthropometric parameters.

\noindent\textbf{Center-of-Mass from Monocular Video.}
A monocular pose detector provides image-space keypoints corresponding to the anatomical joints of the segmental model, but not their metric positions relative to the camera. MuyBridge maps each keypoint through the camera intrinsics to a viewing ray $\mathbf{d}_j$, such that a joint at athlete range $t$ lies at $\mathbf{p}_j=t\,\mathbf{d}_j$. The normalized depth field is sampled at the detected keypoints to provide relative range information. Metric scale is obtained from sex-specific de Leva segment-length proportions scaled by subject stature, with ground contact and ballistic motion providing additional anchors when available. These measurements recover the metric range $t$, after which the resulting joint positions define the segment endpoints in Equation~\ref{eq:segcom} and the anthropometric parameters $\rho_s$ and $m_s$ give the whole-body CoM through Equation~\ref{eq:bodycom}.
 
\subsection{System Overview}
\label{sec:overview}
The goal of MuyBridge is to estimate metric center-of-mass trajectories of athletes directly from monocular phone video. MuyBridge (Figure~\ref{fig:pipeline}) consists of three stages: (i) a compact 2D pose network that predicts anatomical keypoints, (ii) a single-step monocular depth network that provides normalized depth at those locations, and (iii) an analytic metric-fusion stage that combines relative depth with anthropometric and physical range cues to recover metric joint positions and compute whole-body CoM.

\subsection{2D Pose Network}
\label{sec:posebranch}
The pose branch estimates the 26 anatomical keypoints of the Halpe-26~\cite{fang2023alphapose} layout from each RGB frame. We use a compact
RTMPose-style architecture~\cite{jiang2023rtmpose} with an
RTMDet person detector~\cite{lyu2022rtmdet}, a CSPNeXt backbone
~\cite{chen2024cspnext}, a gated attention unit, and a SimCC regression
head~\cite{li2022simcc}. The detected athlete is cropped to a
$256 \times 192$ input, and SimCC predicts separate horizontal and vertical
coordinate distributions for each joint. The resulting output is $K=\{(\hat{u}_j,\hat{v}_j,w_j)\}_{j=1}^{26}$, where $(\hat{u}_j,\hat{v}_j)$ is the image-space location of joint $j$ and $w_j$ its confidence. Coordinates are mapped back to the original image before metric fusion. We train and evaluate the pose network on COCO-WholeBody~\cite{jin2020wholebody} and HICO-DET~\cite{chao2018hicodet}, with annotations converted to the Halpe-26 layout with no AthletePose3D or sport-specific supervision. The architecture is subsequently restructured, pruned, and quantized for mobile deployment, described in Section~\ref{sec:pose_deploy}.

\subsection{Monocular Depth Network}
\label{sec:depthbranch}
To provide range information for the detected body keypoints, the depth branch processes the same RGB frame as the pose network and predicts a dense normalized depth field $Z(u,v)$. We use a Marigold-based latent depth model~\cite{ke2025marigold}, consisting of a VAE encoder, a denoising UNet, and a VAE decoder. Rather than iterative diffusion sampling, we apply latent consistency distillation~\cite{luo2023lcm} to reduce inference to a single UNet evaluation. The predicted depth is affine-invariant, so individual keypoint samples provide local relative depth and their aggregate estimates athlete range, while absolute metric scale is resolved during fusion by combining these keypoint-depth measurements with the corresponding anatomical and physical anchors. The network is trained on approximately 1.2M synthetic RGB-D pairs from Hypersim~\cite{roberts2021hypersim} and Virtual KITTI~2~\cite{cabon2020virtualkitti2}, with no AthletePose3D or sport-specific supervision. The architecture is subsequently restructured, pruned, and quantized for mobile execution as described in Section~\ref{sec:depth_deploy}.

\subsection{Metric Center-of-Mass Estimation}
\label{sec:fusion}
After pose and depth estimation, MuyBridge converts image-space observations into metric joint positions for the segmental model. Keypoints are back-projected through the camera intrinsics and paired with depth samples at the same anatomical locations, then fused with anthropometric and physical range cues to recover athlete depth for segmental CoM computation.

\noindent\textbf{Sparse Keypoint-Depth Measurements.}
For frame $f$, keypoint $(u_{f,j},v_{f,j})$ defines the camera ray
\begin{equation}
\mathbf{d}_{f,j}\propto
\mathbf{K}^{-1}[u_{f,j},v_{f,j},1]^\top,
\label{eq:camera_ray}
\end{equation}
and the normalized depth field is sampled at the same anatomical location as
$z_{f,j}=Z_f(u_{f,j},v_{f,j})$. Pose confidence and local depth consistency are used jointly to reject observations inconsistent with the athlete. The retained samples are summarized as
$\tilde z_f=\operatorname{median}_{j\in\mathcal{V}_f}z_{f,j}$,
providing a robust relative range signal. Since $Z_f$ is affine-invariant, $\tilde z_f$ captures changes in athlete range but carries no metric scale. 

\noindent\textbf{Metric Range Estimation.}
Metric range is first estimated from anthropometric segment lengths. For segment $k$, a known length $L_k$ from the de Leva body model, scaled by subject stature or replaced by a measured length when available, and endpoint rays $\mathbf{d}_{f,i}$ and $\mathbf{d}_{f,j}$ give
\begin{equation}
t^{\mathrm{size}}_{f,k} =
\frac{L_k}
{\|\mathbf{d}_{f,i}-\mathbf{d}_{f,j}\|}.
\label{eq:sizecue}
\end{equation}
Because foreshortening biases individual estimates upward, range estimates across visible segments are aggregated using a fixed low quantile. Physical cues provide additional metric observations when available. Ground contact yields range through ray--ground intersection, while airborne motion yields range by relating image-space acceleration to gravity. These observations also anchor the affine-invariant depth signal to metric scale. For frames $\mathcal{A}$ where a physical range observation is available, we estimate
\begin{equation}
(\alpha,\beta)
=
\arg\min_{\alpha,\beta}
\sum_{f\in\mathcal{A}}
w_f
\left(
\alpha\tilde z_f+\beta-t^{\mathrm{anchor}}_f
\right)^2,
\label{eq:depthcal}
\end{equation}
where $t_f^{\mathrm{anchor}}$ is the metric range obtained from ground contact or ballistic motion and $w_f$ is its measurement weight. This gives the depth-derived range
$t^{\mathrm{depth}}_f=\alpha\tilde z_f+\beta$.
The anthropometric, physical, and depth-derived measurements are fused with a constant-velocity range model to obtain the final athlete range $\hat t_f$.

\noindent\textbf{Metric Reconstruction and Segmental CoM.}
The fused athlete range $\hat t_f$ places the predicted keypoint rays in metric camera coordinates. The hip-center ray defines a vertical reconstruction plane at this range, and each keypoint ray is intersected with the plane to obtain metric joint positions $\hat{\mathbf{p}}_{f,j}$. The reconstructed Halpe-26 joint locations are then mapped to the corresponding de Leva segment coordinates for the head, trunk, bilateral thighs, shanks, feet, upper arms, and forearms, with the trunk defined from neck to hip center and hand mass folded into the forearms. The sex-specific $m_s$ and $\rho_s$ parameters are applied to segment endpoints to compute segment centers and whole-body CoM through Equations~\ref{eq:segcom} and~\ref{eq:bodycom}.

\section{On-Device Deployment}
\label{sec:deployment}

\subsection{Pose Network Compression}
\label{sec:pose_deploy}
We restructure the pose network of Section~\ref{sec:posebranch} for efficient execution on the Apple Neural Engine, then apply structured pruning and quantization-aware training.

\noindent\textbf{Architecture restructuring.}
We replace SiLU with ReLU, express the projections in the gated attention unit as convolution--normalization--activation blocks, and replace hard-sigmoid gates with standard sigmoid operations to map the graph to accelerator-supported operators. Pooling and convolution kernels are capped at $7\times7$, except for the large-kernel SimCC block required for coordinate resolution. We further compress the CSPNeXt backbone to width$\times0.275$ and depth$\times0.157$, with channel dimensions rounded to multiples of 8 or 16 for Neural Engine tensor packing. The reduced feature dimensions also lower SRAM pressure and intermediate feature transfers.

\noindent\textbf{Structured pruning.}
We apply GroupFisher structured pruning~\cite{liu2021groupfisher}, concentrating removal in the early network rather than shrinking all stages uniformly. CSPNeXt Stages~0--1 are reduced by 55--65\%, Stages~2--3 by 40--50\%, and Stage~4 by at most 10\%, reducing the network from 3336 to 2389 channels and approximately halving its floating-point computation. The pruned network is then fine-tuned in full precision before quantization.

\noindent\textbf{Mixed-precision quantization.}
We perform quantization-aware training with symmetric per-channel quantization for weights and asymmetric per-tensor quantization for activations. The SimCC softmax remains in floating point because normalization over coordinate bins is sensitive to eight-bit reduction, while the remaining compatible operators are quantized for Neural Engine execution.

\begin{table}[h!]
\centering
{\footnotesize
\begin{tabular}{lccc}
\toprule
Model & FPS $\uparrow$ & AUC $\uparrow$ & PCK@0.1 $\uparrow$ \\
\midrule
Ours (INT8) & \textbf{224} & \textbf{0.902} & \textbf{73.32\%} \\
RTMPose (F32) & 123 & 0.870 & 66.35\% \\
\bottomrule
\end{tabular}}
\caption{Pose accuracy and server-side throughput of the compressed INT8 model and full-size RTMPose reference. AUC is the area under the PCK curve, and PCK@0.1 reports the percentage of keypoints within 10\% of the reference scale.}
\label{tab:pose}
\end{table}
\noindent The compressed model reaches 1.82$\times$ the RTMPose throughput with a 7.0-point PCK@0.1 gain. The accuracy difference reflects Halpe-26 retraining rather than quantization (Table~\ref{tab:pose}).

\subsection{Depth Network Compression}
\label{sec:depth_deploy}
We restructure the depth network of Section~\ref{sec:depthbranch} for efficient mobile execution, applying conditioning removal, operator-compatible restructuring, and mixed-precision quantization.

\noindent\textbf{Conditioning removal.}
Marigold inherits the CLIP text-conditioning of its Stable Diffusion~2 backbone. We found that replacing the text encoder with a fixed learned embedding does not measurably degrade depth performance while removing approximately 340~M parameters.

\noindent\textbf{Operator-compatible restructuring.}
We restructure the remaining network for Core~ML execution. GELU and SiLU activations are replaced with ReLU and Hardswish so that adjacent convolution, normalization, and activation operations can be fused across accelerator tiles. Group and layer normalization are replaced with a batch-adaptive normalization wrapper that yields a static computation graph and can be folded into neighboring convolutions at export. 

\noindent\textbf{Mixed-precision quantization.}
We found that post-training quantization degrades the UNet's predicted depth structure, whereas the frozen VAE maintains stable activation ranges under post-hoc calibration. We therefore apply quantization-aware training to the UNet and post-training quantization to the VAE. Convolution and projection layers use eight-bit symmetric per-channel weight quantization and asymmetric per-tensor activation quantization, while attention, query/key/value projections, softmax, and residual normalization remain in FP16 because full INT8 attention degrades depth accuracy.

\begin{table}[h!]
\centering
{\footnotesize
\setlength{\tabcolsep}{3.5pt}
\resizebox{\columnwidth}{!}{%
\begin{tabular}{llccc}
\toprule
Dataset & Model & AbsRel $\downarrow$ & RMSE$_{\text{lin}}$ $\downarrow$ & $\delta_1$ $\uparrow$ \\
\midrule
KITTI-Depth~\cite{uhrig2017sparsity}            & Marigold (F32) & 0.132 & 3.588 & 0.852 \\
                        & Ours (INT8)        & 0.155 & 3.885 & 0.776 \\
\midrule
Hypersim$^{\dagger}$~\cite{roberts2021hypersim}                & Marigold (F32) & 0.128 & 0.738 & 0.880 \\
                        & Ours (INT8)        & 0.149 & 0.800 & 0.849 \\
\midrule
NYU-Depth~\cite{NYUdepthsilberman2012indoor}               & Marigold (F32) & 0.053 & 0.230 & 0.976 \\
                        & Ours (INT8)        & 0.061 & 0.253 & 0.965 \\
\bottomrule
\end{tabular}%
}}
\caption{Depth accuracy compressed network compared with
full-precision Marigold reference. AbsRel, linear RMSE, $\delta_1$, $^{\dagger}$held-out test split, all follow standard monocular depth evaluation.}
\label{tab:depth}
\end{table}
\noindent  Quantization introduces a modest accuracy loss across datasets, with AbsRel increasing by 0.008--0.023 and $\delta_1$ decreasing by 0.011--0.076 (Table~\ref{tab:depth}). The largest degradation occurs on KITTI-Depth, while NYU-Depth remains comparatively stable.

\subsection{Pipeline Deployment and Profiling}
\label{sec:pipeline_deploy}
\label{sec:profiling}
Both networks are converted from PyTorch to Core~ML and compiled into fixed-shape \texttt{.mlmodelc} binaries. The depth model is split into encoder, denoiser, and decoder graphs to avoid poor accelerator mapping of a single monolithic graph. MuyBridge runs as a native Swift application on an iPhone~15, with AVFoundation supplying camera frames, Core~ML executing the perception networks, and CPU-based metric fusion and CoM computation.

\noindent\textbf{Setup.}
Camera intrinsics are obtained directly from the device, while a one-time scene calibration provides the ground plane and landmark offsets used by the physical range cues of Section~\ref{sec:fusion}. Subject sex and stature are specified once before recording, with stature scaling the anthropometric segment lengths and sex selecting the corresponding de Leva parameters.

\noindent\textbf{Runtime.}
The pose network sustains 63~FPS, whereas the depth field refreshes at 2.86~Hz. MuyBridge therefore performs fusion at the pose rate using the most recent depth field. This asynchronous schedule matches the role of each branch, pose captures frame-to-frame articulation, while depth supplies a more slowly varying range signal and local consistency measurements. Output begins after the first depth field is available, with the anatomical range cue providing metric range from the first usable frame.

\noindent\textbf{Profiling.}
We profile the compiled pipeline on an iPhone~15 over 100 forward passes. Latency and throughput are measured from the deployed Core~ML models, energy from the Xcode energy log, peak memory from Instruments, and model size from the compiled \texttt{.mlmodelc} bundles.

\begin{table}[h!]
\centering
{\footnotesize
\begin{tabular}{@{}lrrrr@{}}
\toprule
Model & Lat. (ms) $\downarrow$ & FPS $\uparrow$ & Size (MB) & Energy (mJ) $\downarrow$ \\
\midrule
Pose network & 15.58 & 63.0 & 17.4  & 23.4 \\
Depth network & 349.9 & 2.86 & 928.9 & 615.0 \\
\midrule
Full$^{\dagger}$, $256\times192$ & 349.9 & 2.86 & 946.3 & 638.4 \\
Full, $256\times256$ & 424.6 & 2.35 & 946.3 & 752.4 \\
Full, $320\times320$ & 384.7 & 2.60 & 946.3 & 690.4 \\
\bottomrule
\end{tabular}}
\caption{On-device profiling on an iPhone~15. Component rows report the deployed $256\times192$ configuration, while full-pipeline rows compare $^{\dagger}$MuyBridge input resolutions.}
\label{tab:runtime}
\end{table}
\noindent At the deployed resolution, pose inference requires 15.58~ms while a depth update requires 349.9~ms, making the depth branch the dominant cost in model size, energy, and memory (Table~\ref{tab:runtime}). A full pass costs 349.9\,ms and 638\,mJ, which is the cold-start configuration rather than steady-state operation. Across resolutions, latency is non-monotonic, with $320\times320$ executing faster than $256\times256$, due to accelerator-specific tiling and core allocation. Peak memory reaches 1{,}185~MB for the complete pipeline, compared with 126.8~MB for pose alone. Per-submodule profiling, tail latencies, and compute are reported in the appendix.

\section{Experiments}
\label{sec:experiments}
\subsection{Datasets and Evaluation Protocol}
\label{sec:protocol}
We evaluate MuyBridge on AthletePose3D~\cite{yeung2025athletepose3d}, which contains eight athletes and approximately 1.3M synchronized multi-view frames. We partition the evaluation into running, track and field, and figure skating to characterize motion-dependent performance using test partitions comprising 1{,}028 sequence-camera pairs and 235{,}183 frames. Neither perception network is trained or fine-tuned on AthletePose3D. For each sequence, camera intrinsics and the one-time scene calibration provide the geometric inputs required by Section~\ref{sec:fusion}. Subject sex selects the corresponding de Leva~\cite{deleva1996adjusted} anthropometric parameters, while stature scales sex-specific segment-length proportions used for metric range estimation. Reference trajectories are used only to compute evaluation targets and are not used during inference. Reference CoM is computed from the provided 3D joints using the segmental model of Section~\ref{sec:basis} and the same anthropometric formulation as our predictions. We report 3D CoM MAE and its lateral, vertical, and depth components and range AbsRel, $\mathrm{AbsRel}=\frac{1}{T}\sum_t \frac{\lvert \hat{Z}_t-Z_t\rvert}{Z_t}$. Errors are averaged over frames within each sequence-camera pair and reported as medians over pairs.
 
\subsection{Center-of-Mass Accuracy}
\label{sec:sysperf}
We first characterize overall CoM accuracy across motion types, reporting lateral ($X$), vertical ($Y$), and camera-depth ($Z$) errors to determine how metric localization varies across movement regimes.

\begin{table}[h!]
\centering
{\footnotesize
\setlength{\tabcolsep}{3.2pt}
\begin{tabular}{lcccccc}
\toprule
& \multicolumn{3}{c}{Metric localization}
& \multicolumn{3}{c}{Axis MAE (mm)} \\
\cmidrule(lr){2-4}\cmidrule(lr){5-7}
Regime & Range (m) & 3D (mm) & AbsRel (\%) & $X$ & $Y$ & $Z$ \\
\midrule
Running        & 4.4  & 187 & 3.6 & 44  & \textbf{33} & 166 \\
Track \& field & 5.4  & 185 & 2.3 & 94  & 39          & 117 \\
Skating        & 10.1 & 707 & 6.6 & 167 & 41          & 672 \\
\bottomrule
\end{tabular}}
\caption{Metric CoM accuracy across motion regimes. Range is mean athlete distance from the camera; 3D and per-axis errors are MAE in mm. Values are medians over sequence-camera pairs after averaging within each pair.}
\label{tab:headline}
\end{table}
\noindent The larger 3D errors are concentrated along camera depth, which contributes 166 of 187~mm in running, 117 of 185~mm in track and field, and 672 of 707~mm in figure skating (Table~\ref{tab:headline}). Lateral and vertical errors remain substantially smaller, indicating that monocular range recovery is the dominant source of metric localization error. We next examine how camera geometry and motion affect this range error.

\subsection{Comparison with Recent Methods}
\label{sec:sota_comparison}
We compare MuyBridge with recent monocular human reconstruction and 3D pose methods using their official pretrained models without fine-tuning on AthletePose3D. Predictions are converted to whole-body CoM using the same de Leva formulation and evaluated on the same test partitions and sequence--camera aggregation protocol. Absolute evaluation uses native camera-space predictions when available, while root-relative evaluation removes pelvis translation from the prediction and reference.

\begin{table}[t]
\centering
{\footnotesize
\setlength{\tabcolsep}{3.2pt}
\begin{tabular}{lcccc}
\toprule
Method & Running & Track \& field & Skating & Mobile \\
\midrule

\multicolumn{5}{l}{\itshape Absolute 3D CoM MAE (mm)} \\
MeTRAbs-S~\cite{sarandi2021metrabs}
    & 224 & \textbf{120} & \textbf{231} & \xmark \\
HMR2.0~\cite{goel2023humans4d}
    & N/A & N/A & N/A & \xmark \\
CameraHMR~\cite{patel2025camerahmr}
    & 392 & \underline{180} & \underline{534} & \xmark \\
NLF~\cite{sarandi2024nlf}
    & \textbf{52} & 209 & 646 & \xmark \\
MuyBridge
    & \underline{187} & 185 & 707 & \cmark \\

\midrule

\multicolumn{5}{l}{\itshape Root-relative 3D CoM MAE (mm)} \\
MeTRAbs-S~\cite{sarandi2021metrabs}
    & 67 & \underline{72} & \underline{57} & \xmark \\
HMR2.0~\cite{goel2023humans4d}
    & 70 & 80 & 69 & \xmark \\
CameraHMR~\cite{patel2025camerahmr}
    & 68 & 74 & 58 & \xmark \\
NLF~\cite{sarandi2024nlf}
    & \underline{64} & \textbf{68} & \textbf{53} & \xmark \\
MotionAGFormer~\cite{mehraban2024motionagformer}
    & \textbf{45} & 79 & 90 & \xmark \\
MuyBridge
    & \textbf{45} & 118 & 76 & \cmark \\

\bottomrule
\end{tabular}
}
\caption{Comparison with recent monocular reconstruction and 3D pose methods on AthletePose3D. Values are median 3D CoM (mm) MAE, with best and second-best results shown in \textbf{bold} and \underline{underline}. Mobile indicates demonstrated mobile deployment in the cited work. HMR2.0 does not provide calibrated metric camera-space placement, and MotionAGFormer is provided per-frame ground-truth scale due to its scale-ambiguous predictions.}
\label{tab:sota_comparison}
\end{table}
Dedicated reconstruction methods achieve lower absolute CoM error in track and field and skating, as well as lower root-relative error in track and field (Table~\ref{tab:sota_comparison}). MuyBridge ranks second in absolute running error and matches MotionAGFormer at 45~mm root-relative error, although MotionAGFormer receives per-frame ground-truth scale. Unlike these reconstruction-specific baselines, MuyBridge uses generic pose and depth supervision with an explicit segmental biomechanical model incorporating anthropometric and physical constraints. This compact formulation enables end-to-end execution on an iPhone~15, making MuyBridge the only evaluated method with demonstrated mobile deployment.

\subsection{Analysis Across Motion Regimes}
\label{sec:regimes}
\noindent\textbf{Cyclic locomotion.}
Running provides a baseline athletic motion regime with repeated acceleration and deceleration, periodic vertical and lateral CoM motion, and sustained translation through the scene. Across the running partition, MuyBridge reaches 187~mm 3D CoM MAE and 3.6\% range AbsRel at a mean athlete range of 4.4~m.

\begin{figure}[h!]
\centering
\includegraphics[width=\columnwidth]{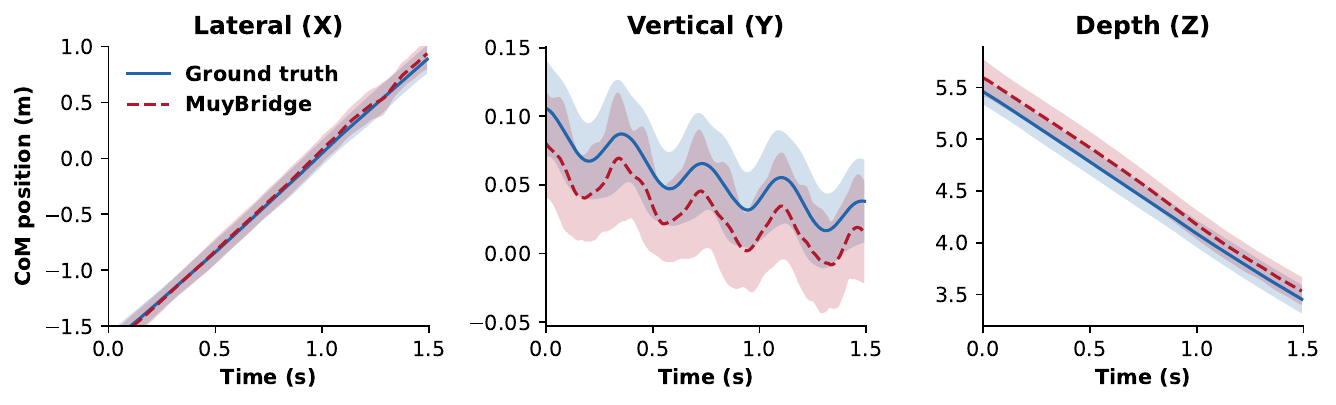}
\caption{Running CoM trajectories from a representative camera view, averaged across 37 sequences. Shading denotes one standard deviation across pairs.}
\label{fig:running}
\end{figure}
\noindent The lateral and vertical CoM trajectories closely follow the reference, while depth retains a systematic offset during forward translation (Figure~\ref{fig:running}). Across all running sequences and camera views, median per-axis error is 44~mm lateral, 33~mm vertical, and 166~mm in depth, making camera-to-athlete range the dominant source of 3D error.

\noindent\textbf{Confined rotational movement.}
Track and field introduces larger limb excursions and rapid changes in body configuration than running, making whole-body CoM more sensitive to the recovery of individual segment positions. The throwing events also involve substantial body rotation while remaining largely confined spatially. MuyBridge reaches 185~mm 3D CoM MAE and 2.3\% range AbsRel at a mean range of 5.4~m.
 
\begin{figure}[h!]
\centering
\includegraphics[width=\columnwidth]{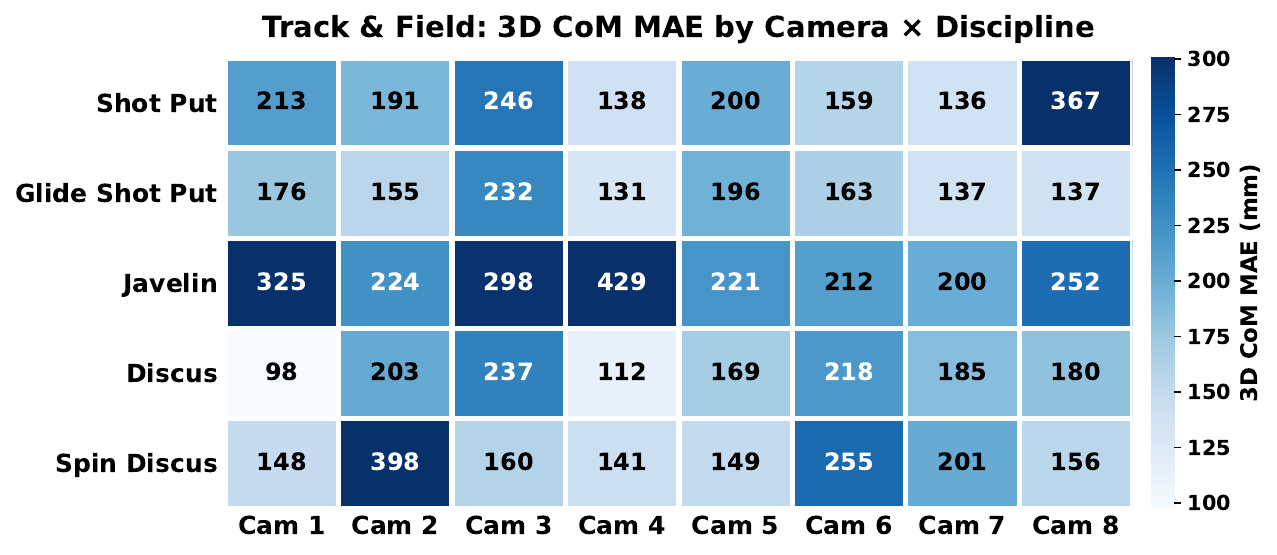}
\caption{3D CoM MAE across camera views and throwing disciplines. Cameras 1--8 denote the eight synchronized AthletePose3D camera views used for evaluation.}
\label{fig:tf_heatmap}
\end{figure}

\noindent There is a clear distinction between confined throwing motions and javelin. Across the fixed-circle events, 21 of 32 camera--discipline combinations remain below 200~mm 3D CoM MAE (Figure~\ref{fig:tf_heatmap}). Javelin instead ranges from 200 to 429~mm across cameras. Unlike the other events, javelin combines large segmental motion with a run-up and sustained translation through camera depth, increasing the difficulty of metric range recovery.

\begin{figure}[h!]
\centering
\includegraphics[width=\columnwidth]{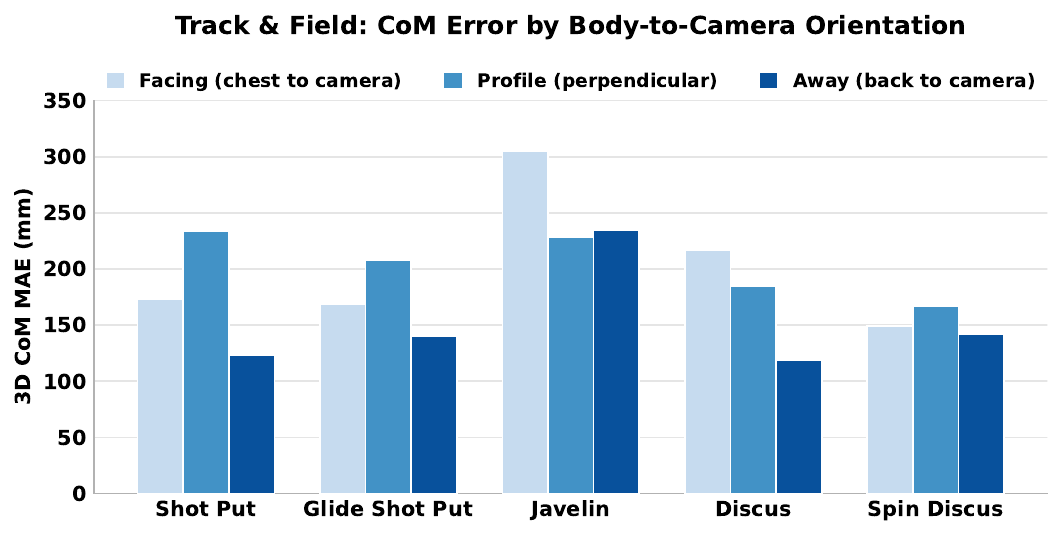}
\caption{3D CoM MAE stratified by body-to-camera orientation. Each frame is independently assigned to a facing, profile, or away orientation bin based on GT pelvis heading, so a sequence may contribute to multiple bins.}
\label{fig:tf_orientation}
\end{figure}

\noindent We stratify frames by body-to-camera orientation using the reference hip axis only for analysis (Figure~\ref{fig:tf_orientation}). Back-facing configurations yield the lowest error in four of five events, while profile views exceed facing views in three of five. Perpendicular orientations can increase overlap and occlusion of distal joints, reducing the landmarks available for anatomical range estimation and depth sampling. Javelin is the exception, where translational motion during the run-up dominates the orientation effect.

\noindent\textbf{Ballistic Movement.}
\label{sec:ballistic}
Figure skating provides the most challenging regime, combining repeated airborne motion with an athlete range of 10.1~m, nearly twice the other motion regimes. Jumps introduce rapid vertical acceleration, large changes in body configuration, and extended periods without ground contact. MuyBridge reaches 707~mm 3D CoM MAE and 6.6\% range AbsRel in this setting.

\begin{figure}[h!]
\centering
\includegraphics[width=\columnwidth]{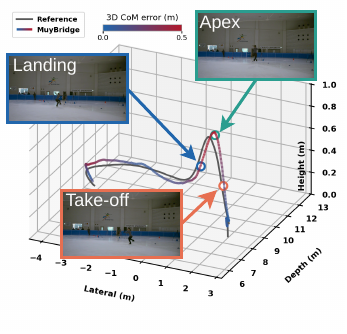}
\caption{Recovered metric CoM trajectory for a representative Toeloop recorded from approximately 10~m. The height axis is exaggerated for visibility.}
\label{fig:skate_traj}
\end{figure}

\noindent MuyBridge recovers the overall spatial trajectory of the jump while the athlete moves several meters through camera depth (Figure~\ref{fig:skate_traj}). At this range, small image-space perturbations produce larger metric displacements, and airborne motion removes the ground-contact cue used for range estimation. We therefore separate grounded and airborne phases to determine how strongly flight contributes to the skating error.

\begin{figure}[h!]
\centering
\includegraphics[width=\columnwidth]{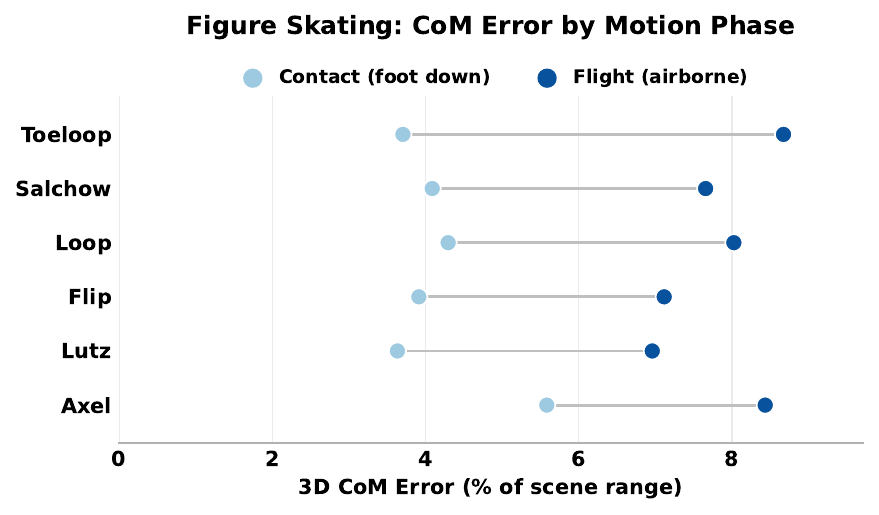}
\caption{3D CoM error during contact and flight across six figure-skating jump types. Each point shows the phase-aggregated error for one jump type, normalized by mean athlete-to-camera range, gray lines connect contact and flight values for the same jump.}
\label{fig:skate}
\end{figure}

\noindent We separate contact and flight phases to characterize performance during airborne ballistic motion (Figure~\ref{fig:skate}). Contact error ranges from 3.6--5.6\% of athlete range, compared with 7.0--8.7\% during flight, corresponding to a 1.5--2.3$\times$ increase. The same pattern appears across all six jump types, indicating that the degradation is associated with airborne motion rather than a specific jump.

\subsection{Ablation Study}
\label{sec:ablation}
\noindent\textbf{Representation and range cues.}
We ablate the key components of MuyBridge to quantify the contributions of keypoint-based depth sampling, the segmental body representation, and individual metric range cues.

\begin{table}[h!]
\centering
{\footnotesize
\setlength{\tabcolsep}{4pt}
\begin{tabular}{lccc}
\toprule
Variant & Running & Track \& field & Skating \\
\midrule
Root-relative CoM$^{*}$         & 45  & 118  & 76 \\
Constant sequence range         & 721 & 235 & 1890 \\
Torso-only body model           & 284 & 252 & 733 \\
MuyBridge                       & \textbf{187} & \textbf{185} & \textbf{707} \\
\midrule
No ground-contact cue           & 355 & 808 & 1220 \\
No keypoint-depth cue           & 387 & 316 & 939 \\
\bottomrule
\end{tabular}}
\caption{Ablation of the metric CoM formulation. Values are median 3D CoM MAE in mm. The upper block evaluates alternative range and body representations, while the lower block removes individual range cues. $^{*}$Root-relative CoM excludes absolute camera-to-athlete translation and is therefore not directly comparable to the metric variants.}
\label{tab:ablation}
\end{table}
\noindent The ablations confirm that each component of the metric formulation contributes to final CoM accuracy (Table~\ref{tab:ablation}). Holding athlete range constant increases error from 187 to 721~mm in running and from 707 to 1890~mm in skating, while restricting reconstruction to the torso increases error to 284, 252, and 733~mm across the three motion regimes. Removing either the ground-contact or keypoint-depth cue also increases error across all motions. The full MuyBridge formulation achieves the lowest metric error in every case, showing that sparse depth, physical range cues, and segmental anatomy provide complementary information for metric CoM estimation.

\noindent\textbf{Sparse depth sampling.}
We vary the anatomical keypoints used to sample the relative depth field and evaluate their effect on the resulting range signal and metric CoM accuracy.

\begin{table}[h!]
\centering
{\footnotesize
\setlength{\tabcolsep}{4pt}
\begin{tabular}{lcccccc}
\toprule
 & \multicolumn{2}{c}{Running} & \multicolumn{2}{c}{Track \& field} & \multicolumn{2}{c}{Skating} \\
\cmidrule(lr){2-3}\cmidrule(lr){4-5}\cmidrule(lr){6-7}
Depth samples & $r$ & mm & $r$ & mm & $r$ & mm \\
\midrule
All 26 keypoints  & \textbf{0.949} & \textbf{187} & \textbf{0.779} & \textbf{185} & \textbf{0.642} & 707 \\
12 limb keypoints & 0.929 & 342 & 0.201 & 186 & 0.595 & 706 \\
6 torso keypoints & 0.906 & 341 & 0.313 & \textbf{185} & 0.590 & \textbf{633} \\
Hip only          & 0.933 & 344 & 0.261 & 187 & 0.572 & 643 \\
\bottomrule
\end{tabular}}
\caption{Effect of anatomical keypoint sampling on metric CoM estimation. $r$ is the correlation between sampled relative depth and true athlete range, and mm is median 3D CoM MAE. All 26 keypoints form the default MuyBridge configuration.}
\label{tab:sampling}
\end{table}
\noindent All 26 keypoints provide the strongest range correlation in every regime and reduce running error from 341--344~mm to 187~mm (Table~\ref{tab:sampling}). Track and field is comparatively insensitive to the sampling set, remaining between 185 and 187~mm. In skating, torso-only sampling achieves the lowest error at 633~mm despite lower range correlation than the full set, showing that correlation with athlete range does not alone determine final CoM accuracy. Overall, broad keypoint coverage provides the most consistent performance across motion regimes.

\noindent\textbf{Depth usage.}
After ablating the keypoint-depth cue as a whole, we separate its two uses in metric fusion. Sparse depth provides an aggregated depth-derived range estimate and filters anatomical samples whose depth is inconsistent with the athlete.

\begin{table}[h!]
\centering
{\footnotesize
\setlength{\tabcolsep}{4pt}
\begin{tabular}{lccc}
\toprule
Variant & Running & Track \& field & Skating \\
\midrule
MuyBridge                         & \textbf{187} & \textbf{185} & \textbf{707} \\
No depth-derived range estimate   & 392 & 207 & 904 \\
No depth-consistency filtering    & 351 & 192 & 730 \\
\bottomrule
\end{tabular}}
\caption{Median 3D CoM MAE in mm after separately ablating the two uses of sparse depth. Removing the depth-derived range estimate retains depth-consistency filtering, while removing consistency filtering retains the depth-derived range estimate.}
\label{tab:depth_ablation}
\end{table}

\noindent Depth-derived range estimate provides the larger overall contribution, with its removal increasing error by 205~mm in running and 197~mm in skating (Table~\ref{tab:depth_ablation}). Depth-consistency filtering also reduces running error by 164~mm, with smaller effects in track and field and skating. Using both gives the lowest error across all three regimes.

\section{Conclusion}
We presented MuyBridge, a mobile system for metric center-of-mass estimation from monocular video using a three-stage fusion method based on keypoint depth, anatomical constraints, and physical range cues. Across running, track and field, and figure skating, MuyBridge maintains vertical CoM error between 33 and 41~mm and localizes athlete range to 2.3--6.6\% AbsRel while sustaining CoM output at the 63~FPS pose-estimation rate with asynchronous 2.86~Hz depth updates on an iPhone~15. Among the evaluated methods, MuyBridge remains competitive for running while being the only system demonstrated on mobile hardware. Ablations further show that segmental anatomy, sparse depth, dynamic range estimation, and physical range cues each contribute to final metric CoM accuracy. Absolute localization remains primarily limited by camera-to-athlete range, particularly under sustained translation and long-range airborne motion. MuyBridge also assumes a one-time scene calibration, and its anthropometric reconstruction uses stature-scaled, sex-specific population segment proportions. AthletePose3D contains only eight athletes and provides markerless rather than marker-based reference trajectories. Future work on calibration robustness, subject-adaptive anthropometry, motion-adaptive range estimation, and lighter depth backbones could further enhance mobile biomechanics beyond controlled settings.
 
{
    \small
    \bibliographystyle{ieeenat_fullname}
    \bibliography{main}
}

\appendix

\section{Qualitative Reconstruction}
\label{app:qualitative}
\begin{figure*}[h!]
\centering
\includegraphics[width=\linewidth]{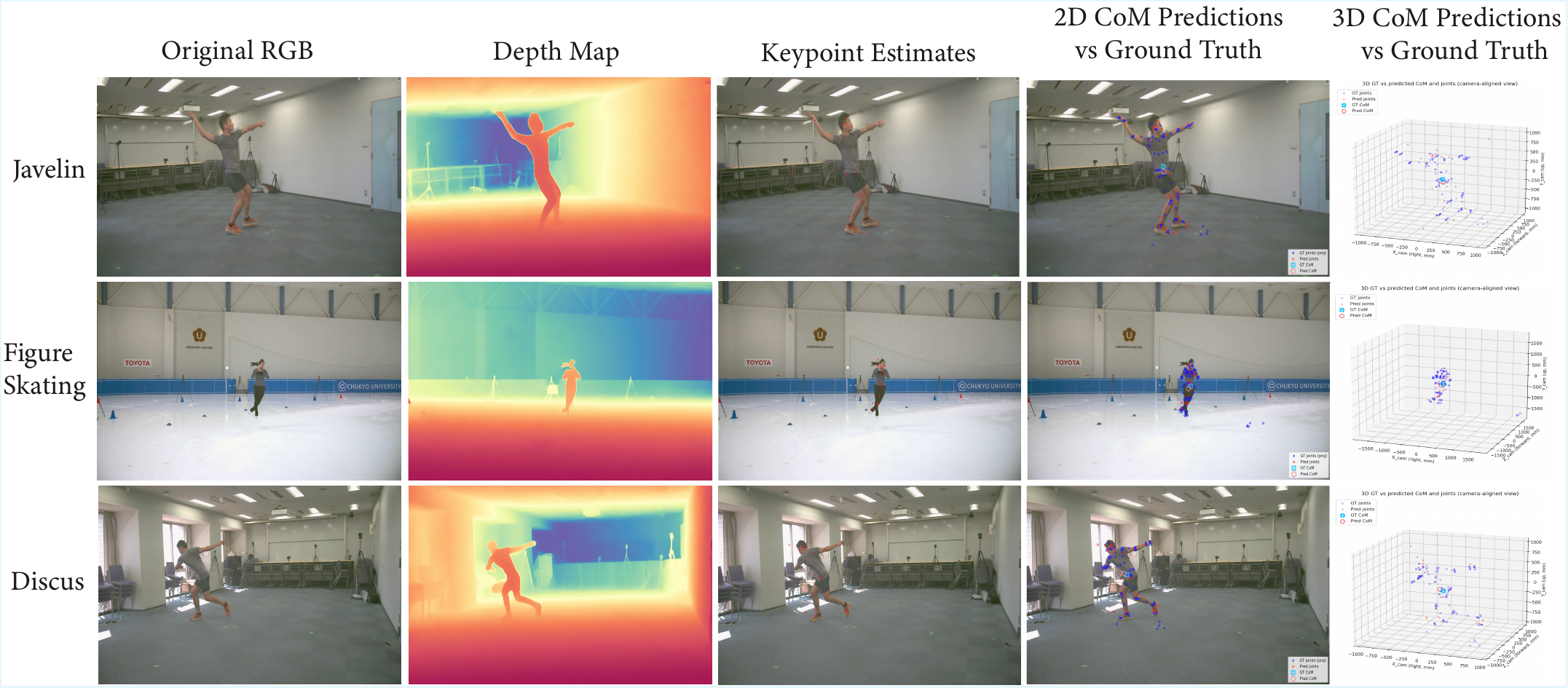}
\caption{Qualitative end-to-end reconstruction on three AthletePose3D sequences. Each row shows the RGB frame, the predicted depth field, the 2D keypoints, the image-plane center of mass against the reference, and the recovered metric center of mass.}
\label{fig:supp_qualitative}
\end{figure*}

\noindent Figure~\ref{fig:supp_qualitative} shows end-to-end output on three
AthletePose3D sequences. The depth field is sampled only at keypoint locations,
for whole-body range and consistency gating.

\section{Extended Pose Network Evaluation}
\label{app:pose-eval}
\noindent Table~\ref{tab:pose-profile-full} attributes the pose cost of
Table~\ref{tab:runtime} to the detector and the keypoint backbone. The backbone
dominates, at 12.53 of 15.58~ms and 100.2 of 126.8~MB.

\begin{table*}[h!]
  \centering
  {\footnotesize
  \begin{tabular}{lcccccc}
    \toprule
    Variant & Latency (ms) $\downarrow$ & FPS $\uparrow$ & Size (MB) & Energy (mJ) $\downarrow$ & Peak mem (MB) & FLOPs / MACs \\
    \midrule
    BBox head         & 3.05  & 328.0 & 3.3  & 4.6  & 26.6  & 0.31 GF / 0.155 GM \\
    Keypoint backbone & 12.53 & 81.0  & 14.1 & 18.8 & 100.2 & 0.67 GF / 0.335 GM \\
    Full HPE network  & 15.58 & 63.0  & 17.4 & 23.4 & 126.8 & 0.98 GF / 0.490 GM \\
    \bottomrule
  \end{tabular}}
  \caption{Per-submodule on-device profiling of the pose network on the Neural Engine.}
  \label{tab:pose-profile-full}
\end{table*}

\noindent Table~\ref{tab:pose-tail-latency} reports mean and percentile
latencies from the same 100-pass trace. The p99 stays within 2.5~ms of the
median, so throughput is stable under continuous execution.

\begin{table*}[h!]
  \centering
  {\footnotesize
  \begin{tabular}{lccccccc}
    \toprule
    Variant & p50 & p90 & p99 & Mean & Energy (mJ) $\downarrow$ & Mem before & Mem peak \\
    \midrule
    BBox head         & 3.12  & 3.57  & 3.79  & 3.05  & 4.6  & 15.58 & 26.56  \\
    Keypoint backbone & 13.36 & 13.89 & 14.23 & 12.53 & 18.8 & 31.61 & 100.25 \\
    Full pass         & 16.50 & 17.50 & 18.00 & 15.58 & 23.4 & 47.20 & 126.80 \\
    \bottomrule
  \end{tabular}}
  \caption{Tail latencies and memory behaviour of the pose network on an iPhone~15. Latencies in ms, memory in MB.}
  \label{tab:pose-tail-latency}
\end{table*}

\section{Extended Depth Network Evaluation}
\label{app:depth-eval}
\noindent Table~\ref{tab:depth-accuracy-full} extends Table~\ref{tab:depth} to
the full monocular-depth metric suite. The ordering of Table~\ref{tab:depth}
holds across every metric.

\begin{table*}[h!]
  \centering
  {\footnotesize
  \begin{tabular}{llcccccccccc}
    \toprule
    Dataset & Model & AbsRel $\downarrow$ & SqRel $\downarrow$ & RMSE$_\text{lin}$ $\downarrow$ & RMSE$_\text{log}$ $\downarrow$ & log10 $\downarrow$ & $\delta_1$ $\uparrow$ & $\delta_2$ $\uparrow$ & $\delta_3$ $\uparrow$ & iRMSE $\downarrow$ & SILog $\downarrow$ \\
    \midrule
    \multirow{2}{*}{KITTI-Depth}
      & Marigold (F32) & 0.132 & 0.574  & 3.588 & 0.2640 & 0.0662 & 0.852 & 0.972 & 0.990 & 823.0  & 26.04 \\
      & Ours (INT8)    & 0.155 & 0.688  & 3.885 & 0.2800 & 0.0738 & 0.776 & 0.952 & 0.985 & 679.1  & 27.70 \\
    \midrule
    \multirow{2}{*}{Hypersim$^{\dagger}$}
      & Marigold (F32) & 0.128 & 0.184  & 0.738 & 0.1680 & 0.0494 & 0.880 & 0.956 & 0.981 & 48.4   & 16.62 \\
      & Ours (INT8)    & 0.149 & 0.226  & 0.800 & 0.2271 & 0.0610 & 0.849 & 0.945 & 0.971 & 46.2   & 22.23 \\
    \midrule
    \multirow{2}{*}{NYU-Depth}
      & Marigold (F32) & 0.053 & 0.0216 & 0.230 & 0.0804 & 0.0226 & 0.976 & 0.993 & 0.998 & 0.0419 & 8.02 \\
      & Ours (INT8)    & 0.061 & 0.0259 & 0.253 & 0.0891 & 0.0259 & 0.965 & 0.993 & 0.998 & 0.0469 & 8.89 \\
    \bottomrule
  \end{tabular}}
  \caption{Full depth accuracy of the compressed network against the full-precision Marigold reference, extending Table~\ref{tab:depth}. Evaluation follows the Marigold protocol; $^{\dagger}$ denotes the held-out Hypersim test split.}
  \label{tab:depth-accuracy-full}
\end{table*}

\noindent Table~\ref{tab:depth-profile-full} attributes the 349.9~ms depth
update to its three graphs. The UNet carries 210.7~ms, 845.0~MB, and the full
1{,}185~MB peak, so it sets the cost of the branch.

\begin{table*}[h!]
  \centering
  {\footnotesize
  \begin{tabular}{llcccccc}
    \toprule
    Module & Resolution & Latency (ms) $\downarrow$ & FPS $\uparrow$ & Size (MB) & Energy (mJ) $\downarrow$ & Peak mem (MB) & FLOPs / MACs \\
    \midrule
    Encoder & 256$\times$192 & 46.68 & 21.4 & 34.3  & 70.0  & 74   & 2.18 GF / 1.09 GM \\
            & 256$\times$256 & 61.26 & 16.3 & 34.3  & 91.9  & 81   & 2.18 GF / 1.09 GM \\
            & 320$\times$320 & 55.67 & 18.0 & 34.3  & 83.5  & 98   & 2.18 GF / 1.09 GM \\
    \midrule
    UNet    & 256$\times$192 & 210.7 & 4.75 & 845.0 & 406.1 & 1185 & 51.99 GF / 25.99 GM \\
            & 256$\times$256 & 242.2 & 4.13 & 845.0 & 455.4 & 1182 & 67.98 GF / 33.99 GM \\
            & 320$\times$320 & 215.6 & 4.64 & 845.0 & 413.3 & 1208 & 103.98 GF / 51.99 GM \\
    \midrule
    Decoder & 256$\times$192 & 92.51 & 10.8 & 49.6  & 138.8 & 66   & 2.20 GF / 1.10 GM \\
            & 256$\times$256 & 121.1 & 8.3  & 49.6  & 181.6 & 66   & 2.20 GF / 1.10 GM \\
            & 320$\times$320 & 113.4 & 9.0  & 49.6  & 170.1 & 67   & 2.20 GF / 1.10 GM \\
    \midrule
    Full pass & 256$\times$192 & 349.9 & 2.86 & 928.9 & 615.0 & 1185 & 56.4 GF / 28.2 GM \\
            & 256$\times$256 & 424.6 & 2.35 & 928.9 & 729.0 & 1182 & 72.4 GF / 36.2 GM \\
            & 320$\times$320 & 384.7 & 2.60 & 928.9 & 667.0 & 1208 & 108.4 GF / 54.2 GM \\
    \bottomrule
  \end{tabular}}
  \caption{On-device profiling of the INT8 depth module, by submodule and input resolution.}
  \label{tab:depth-profile-full}
\end{table*}

\section{Fused Pipeline Profiling}
\label{app:fused-profile}
\noindent Table~\ref{tab:fused-profile-full} adds peak memory and compute to
Table~\ref{tab:runtime}. Peak memory tracks the depth branch across all three
resolutions, since the branches do not peak concurrently.

\begin{table*}[h!]
  \centering
  {\footnotesize
  \begin{tabular}{lcccccc}
    \toprule
    Resolution & Latency (ms) $\downarrow$ & FPS $\uparrow$ & Energy (mJ) $\downarrow$ & Size (MB) & Peak mem (MB) $\downarrow$ & FLOPs / MACs \\
    \midrule
    256$\times$192 & 349.9 & 2.86 & 638.4 & 946.3 & 1185 & 57.4 GF / 28.7 GM \\
    256$\times$256 & 424.6 & 2.35 & 752.4 & 946.3 & 1182 & 73.4 GF / 36.7 GM \\
    320$\times$320 & 384.7 & 2.60 & 690.4 & 946.3 & 1208 & 109.4 GF / 54.7 GM \\
    \bottomrule
  \end{tabular}}
  \caption{Fused pipeline after Core~ML export, extending Table~\ref{tab:runtime} with peak memory and computational cost. Latency reports the critical path under concurrent pose and depth execution, while energy and model size include both branches.}
  \label{tab:fused-profile-full}
\end{table*}

\section{Running Partition by Camera View}
\label{app:running-views}
\noindent Table~\ref{tab:supp_running} breaks the running result of
Table~\ref{tab:headline} down by view. Camera~4 returns the best lateral and
vertical components with the worst depth component, so variation across views
is dominated by camera-depth localization.

\begin{table}[h!]
\centering
{\footnotesize
\setlength{\tabcolsep}{4pt}
\begin{tabular}{lccccc}
\toprule
View & $n$ & 3D CoM & Lateral $X$ & Vertical $Y$ & Depth $Z$ \\
 & & (mm) & (mm) & (mm) & (mm) \\
\midrule
Camera 1 & 30 & 227 & 43 & 33 & 165 \\
Camera 2 & 37 & 216 & 102 & 33 & 181 \\
Camera 3 & 37 & 151 & 53 & 30 & 120 \\
Camera 4 & 37 & 308 & 38 & 21 & 300 \\
\midrule
All views & 141 & 187 & 44 & 33 & 166 \\
\bottomrule
\end{tabular}}
\caption{Running partition by camera view, absolute 3D CoM MAE and its per-axis split, medians over pairs.}
\label{tab:supp_running}
\end{table}

\section{Jump Height in Figure Skating}
\label{app:jumpheight}
\begin{figure}[h!]
\centering
\includegraphics[width=\columnwidth]{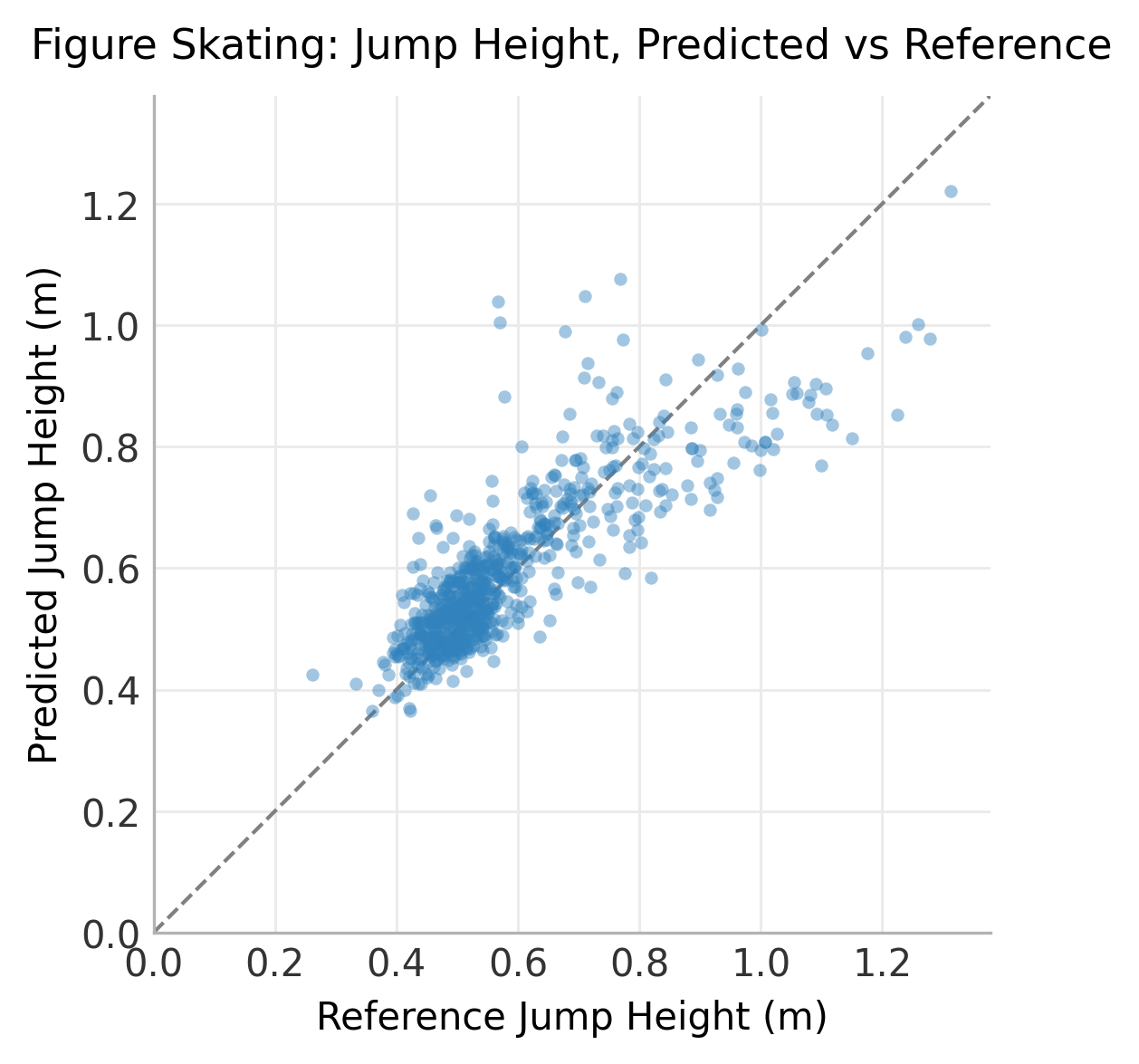}
\caption{Predicted against reference jump height across all figure-skating sequence--camera pairs. The dashed line is the identity.}
\label{fig:jumpheight}
\end{figure}

\noindent Jump height, the takeoff-to-apex rise of the vertical CoM, is
recovered to 62~mm MAE at $r=0.84$ across 719 sequence--camera pairs spanning
0.26--1.31~m reference height, with a mean signed error of $+12$~mm
(Table~\ref{tab:jumpheight}, Figure~\ref{fig:jumpheight}).

\begin{table}[h!]
\centering
{\footnotesize
\setlength{\tabcolsep}{4pt}
\begin{tabular}{lccccc}
\toprule
Jump & $n$ & MAE (mm) $\downarrow$ & $r$ $\uparrow$ & Bias (mm) & GT range (m) \\
\midrule
Toeloop & 120 & 63 & 0.86 & $+16$ & 0.43--1.31 \\
Salchow & 120 & 61 & 0.87 & $+23$ & 0.38--1.02 \\
Loop    & 120 & 67 & 0.89 & $+1$  & 0.26--1.28 \\
Flip    & 120 & 53 & 0.89 & $+13$ & 0.40--1.11 \\
Lutz    & 120 & 74 & 0.67 & $+28$ & 0.47--0.96 \\
Axel    & 119 & 53 & 0.77 & $-9$  & 0.37--1.00 \\
\midrule
\textbf{All} & \textbf{719} & \textbf{62} & \textbf{0.84} & $\mathbf{+12}$ & \textbf{0.26--1.31} \\
\bottomrule
\end{tabular}}
\caption{Jump-height recovery by jump type, computed once per sequence--camera pair as the takeoff-to-apex vertical CoM displacement. Entries report median absolute error over pairs; $r$ is the correlation with reference height, bias is the mean signed error, and GT range is the reference span. This vertical-displacement metric is not directly comparable to the absolute 3D CoM MAE of Table~\ref{tab:headline}.}
\label{tab:jumpheight}
\end{table}

\end{document}